# Uncovering Biases with Reflective Large Language Models

Edward Y. Chang
Computer Science, Stanford University
echang@cs.stanford.edu
February 15$^{th}$, 2024; October 24$^{th}$, 2004 (revised)

## Abstract

Biases and errors in human-labeled data present significant challenges for machine learning, especially in supervised learning reliant on potentially flawed ground truth data. These flaws, including diagnostic errors and societal biases, risk being propagated and amplified through models trained using maximum likelihood estimation. We present the Reflective LLM Dialogue Framework (RLDF), which leverages structured adversarial dialogues between multiple instances of a single LLM or different LLMs to uncover diverse perspectives and correct inconsistencies. By conditioning LLMs to adopt opposing stances, RLDF enables systematic bias detection through conditional statistics, information theory, and divergence metrics. Experiments show RLDF successfully identifies potential biases in public content while exposing limitations in human-labeled data. Our framework supports measurable progress tracking and explainable remediation actions, offering a scalable approach for improving content neutrality through transparent, multi-perspective analysis.

## 1 Introduction

Errors and biases in human-labeled data present critical challenges for machine learning models, especially in healthcare, news, education, and public policy, where their outputs can profoundly shape public perception and decision-making (Mehrabi et al., 2021). Errors, such as diagnostic mistakes, arise from knowledge gaps or lack of expertise, while biases, including ideological and societal distortions, can be consciously or unconsciously introduced by annotators. These flaws compromise the integrity of ground truth data, propagating through machine learning pipelines and generating undesirable outcomes (Kleinberg et al., 2017; Selbst et al., 2019; Baeza-Yates, 2018).

AI systems are particularly vulnerable to these flaws, as models trained on inaccurate or biased ground truth data tend to replicate and amplify these issues through maximum likelihood estimation. In healthcare, diagnostic errors can lead to poor treatment recommendations (Newman-Toker et al., 2023), while in news, partisan annotations—such as labeling a biased article as neutral—mislead both human readers and automated classifiers, distorting public discourse (Mehrabi et al., 2021; Gautam and Srinath, 2024). The impact extends beyond individual sectors: in education, biased data can reinforce stereotypes, while in public policy, it can result in discriminatory decisions. Ensuring that models learn from accurate and impartial ground truth data is therefore essential to the responsible deployment of AI across all domains.

This paper focuses on bias detection and correction in news annotations, using news as the testbed to explore how reflective dialogues among LLMs can mitigate biases. News content is particularly vulnerable to ideological biases, as annotators' personal views often shape the interpretation of politically sensitive topics. Real-world evidence, presented in Section 4, shows how annotation practices differ based on political affiliation. Tables 1 and 2 present real data (Budak et al., 2016) illustrate that Democratic-leaning annotators may judge scandals involving Democrats more harshly than Republicans, and vice versa, highlighting the need for tools to balance these biases.

To address these challenges, we introduce the Reflective LLM Dialogue Framework (RLDF), which implements checks and balances using multiple LLM instances in structured dialogues. RLDF conditions two instances to take opposing stances: one supports the original label, while the other introduces alternative perspectives. These reflective exchanges foster deeper insights and help uncover potential biases, generating more neutral annota-tions through the inclusion of diverse viewpoints. This multi-LLM dialogue approach outperforms the results of a single LLM operating in isolation or providing one-off responses.

RLDF employs conditional statistics, information theory, and divergence metrics to measure the effectiveness of these dialogues. Shannon entropy (Shannon, 1948) quantifies the diversity of perspectives, while mutual information (Cover and Thomas, 2006a) measures the quality of the exchange. To track the convergence toward unbiased outcomes, we apply Jensen-Shannon divergence (JSD) (Lin, 1991), Wasserstein distance (WD) (Kantorovich, 1942), and cross-entropy (CE) (Shore and Johnson, 1980), ensuring that the remediation actions are measurable and transparent for further refinement by human reviewers.

Our empirical studies validate the effectiveness of RLDF, and the contributions of this paper are summarized as follows:

1. Adversarial and Reflective Inspection Framework: RLDF provides a structured framework that encourages adversarial and reflective inspection of ground-truth labels. Through dialogue, participating LLM instances examine, challenge, and explain biases embedded in the original annotations by offering various perspectives. For example, in news annotation, RLDF reveals hidden ideological biases by generating alternative interpretations for politically sensitive content, leading to more neutral labeling.

2. Careful Modulation of Linguistic Behaviors for Balanced Exploration and Exploitation: The effectiveness of RLDF lies in its careful modulation of linguistic behaviors among participating LLM instances, alternating between contentious and conciliatory interactions. This dynamic trade-off fosters exploration of new perspectives while consolidating well-supported viewpoints. Information-theoretic and statistical metrics, including Shannon entropy, mutual information, Jensen-Shannon divergence, Wasserstein distance, and cross-entropy, are employed to measure opinion diversity, information flow, and the strength of the final assessment.

3. Effective Results and Impact on Improving Labels and Mitigating AI Bias: RLDF successfully mitigates AI biases, ensuring more reliable and unbiased model outputs across domains such as news, healthcare (Chang, 2024b), and public policy. These outcomes demonstrate RLDF's significant impact in refining labels, enhancing fairness, and promoting responsible AI deployment.

The remainder of this paper is organized as follows: Section 2 discusses challenges and reviews related work; Section 3 describes the core maxims, theorem, and algorithm; Section 4 presents experiments illustrating successful bias identification and mitigation; and the final section concludes with insights on future work and perceived limitations.

## 2 Related Work

This study focuses on mitigating training data label (ground truth) bias, a primary concern in machine learning (Mehrabi et al., 2021). Accurate labeling is crucial, as a label that aligns with biased content reinforces that bias, while a label that correctly identifies it allows for education and correction (Baeza-Yates, 2018; Danks and London, 2017). This underscores the importance of label accuracy in minimizing bias propagation.

This work specifically addresses mislabeled ground truth and explores remediation actions. Efforts to improve annotation accuracy can be broadly categorized into three approaches:

**Cross-Validation with Multiple Annotators:** Using multiple annotators with statistical aggregation techniques has been shown to reduce individual bias and enhance data reliability (Snow et al., 2008). This method is effective for consensus tasks with clear-cut answers, such as image labeling in ImageNet (Deng et al., 2009; Krizhevsky et al., 2012). However, for more nuanced content like news and Wikipedia articles, majority voting can be problematic. Annotators may possess varying biases on different subjects, and these biases can be unconscious or context-dependent. It is challenging to comprehensively map an annotator's intrinsic tendencies across all possible topics and scenarios. For instance, political affiliation (e.g., Republican or Democrat) does not necessarily predict other beliefs or preferences (such as dietary choices like vegetarianism). Consequently, relying solely on consensus may not effectively mitigate biases, even with a diverse pool of annotators. Moreover, the assumption of a single, absolute truth inherent in human annotation methods can limit the capture of multiple valid viewpoints, particularly in complex or contentious topics (Aroyo and Welty, 2015).

**Cross-Validation between Machine and Human Annotators:** Machine learning models can complement human annotators by enhancing annotation consistency and efficiency ([Wang et al., 2021](#)). Semi-supervised learning methods, exemplified by Snorkel ([Ratner et al., 2017](#)), integrate labeled and unlabeled data to improve model performance. A recent development in this field is the Media Bias Detector (MBD) from the University of Pennsylvania, which utilizes GPT models in conjunction with human raters to analyze potential bias in news articles ([of Pennsylvania, 2024](#)). MBD systematically examines news content from diverse sources, including CNN and Fox News, at regular intervals throughout the day. It employs advanced language models, specifically GPT-3.5 Turbo and GPT-4, to classify articles. The system assigns a political lean score on a scale from -5 (representing strong left-leaning bias) to 5 (indicating strong right-leaning bias). To enhance accuracy, MBD incorporates human verification of the model's outputs.

While MBD attempts to mitigate bias by separating assessments of political lean and tone, it does not explicitly address the inherent biases that may exist within both the GPT models and the human raters. A significant limitation of this approach lies in the fundamental nature LLMs. These models, trained on vast corpora of text data using maximum-likelihood objectives, tend to prioritize statistically prevalent viewpoints. This training methodology can inadvertently lead to the amplification of majority perspectives at the expense of marginalized or less represented viewpoints, potentially introducing subtle but pervasive biases into the analysis.

**Our Contribution: The RLDF Approach** This study aims to address the limitations in MBD and similar frameworks by proposing the Reflective LLM Dialogue Framework (RLDF). RLDF leverages statistical and information-theoretic principles to uncover and balance diverse perspectives, ensuring that both majority and minority viewpoints are adequately represented. Unlike MBD, RLDF introduces structured dialogues between LLM instances, which facilitate deeper reflection and transparent bias mitigation. This approach ensures that annotations are not only accurate but also fair and impartial, improving the reliability of machine learning models across domains.

## 3 Methodology

This section presents our approach in two parts: *debiasing procedure* and *optimization techniques*.

### 3.1 Debiasing Procedure: EVINCE Algorithm

Building upon the theoretical foundations of SocraSynth ([Chang, 2023a](#)), EVINCE (Entropy and Variation in Conditional Exchanges) ([Chang, 2024b](#)) promotes content neutrality through the use of structured dialogues among LLMs. In this section, we describe how we customize EVINCE to perform debiasing effectively.

**Exploring Divergent Viewpoints** Our goal is to generate a broad range of perspectives, even for binary categories such as political leanings. We condition one LLM to support the current ground-truth label while another opposes it, encouraging diverse probability distributions. This approach ensures substantive diversity instead of trivial disagreements. For example, we prefer distributions like (0.5, 0.5) (equal preference on two sub-classes) vs. (1, 0), over mirrored opposites like (1, 0) and (0, 1) (detailed further in Section [3.2](#).).

**Modulating contentiousness** EVINCE dynamically adjusts the intensity of debates using information-theory metrics such as entropy, cross-entropy, and mutual information (see Appendix B). Each LLM generates top-k probability distributions of labels, which EVINCE analyzes to guide subsequent interactions. The contentiousness level is adjusted to either encourage exploration or promote convergence as needed.

In the initial dialogue iterations, we prefer low mutual information and high Wasserstein distance between two LLMs' predictions distributions reflect an explorative phase that encourages divergent viewpoints. As agents exchange well-reasoned arguments, mutual information increases, signaling alignment, while Wasserstein distance decreases, indicating convergence. Once sufficient information exchange occurs, EVINCE reduces contentiousness to foster a more conciliatory atmosphere and guide the agents toward consensus.

**Scrutinizing with Reasonableness** Following the modulation of contentiousness, EVINCE focuses on evaluating the reasonableness of each LLM's arguments. Each LLM presents supporting evidence for its predictions, which is assessed based on logic, coherence, and credibility.

To ensure quality control, EVINCE uses CRIT (Chang, 2023b), a reasonableness evaluation module, to flag weak or unsupported arguments. These flagged arguments are reviewed by human moderators, ensuring that faulty reasoning does not persist in the final outcome. This process balances automated reasoning with human oversight, retaining only those perspectives that survive rigorous scrutiny and ensuring that the resulting consensus reflects well-reasoned, unbiased perspectives.

## 3.2 Optimization and Algorithm Specifications

Figure 1 formally specifies the EVINCE algorithm and associated optimization functions. With all proxy metrics and their pros, cons, and combined strengths comprehensively surveyed in Appendix B, Algorithm 1 formally specifies the key steps of EVINCE. We further address its three optimization problems.

1. *Optimizing Initial Conditions.* Use distinct prompts, randomized seeds, and prior distribution constraints to promote meaningful exploration in the first few rounds.
2. *Optimizing Interaction Dynamics.* Dynamically adjust the debate's intensity using divergence metrics, and Wasserstein distance. Ensure fair turn-taking and filter redundant arguments.
3. *Optimizing Convergence Criteria.* Set clear thresholds for Wasserstein distance, divergence metrics, and passing reasonableness checks through CRIT, to determine when consensus is reached. Use a weighted voting mechanism, with human oversight for ambiguous cases. (Mutual information can be omitted if the joint distribution is not assessable.)

**Problem Statement:** Organize a structured dialogue between two equally competent large language models (LLMs), $LLM_A$ and $LLM_B$, to conduct $t$ rounds. At each round $t$, each model produces a probability distribution, denoted as $P_A^{(t)}$ and $P_B^{(t)}$, over $C$ possible outcomes, accompanied by supporting arguments $R_A^{(t)}$ and $R_B^{(t)}$. The goal is to design an iterative debate process that leverages the structured exchange of arguments to enable the models to converge on an optimal prediction distribution $P^*$ across the $C$ classes.

### 3.2.1 Optimize Initial Condition

The initial phase of the EVINCE algorithm aims to induce *dual entropy* and large Wasserstein Distance (WD) (Kantorovich, 1942; Rubner et al., 2000; Villani, 2008) between the LLM-generated distributions. The large WD requirement is intuitive: we want the two LLMs to present different perspectives. When one LLM is conditioned to take one extreme position and the other the opposite, through integrative debate and gradually decreasing debate intensity (while maintaining reasoning quality), they are expected to reach consensus somewhere between their initial positions.

The Entropy Duality Theory (EDT), however, presents a counter-intuitive insight. EDT posits that optimal information exchange occurs when one agent's distribution has high entropy (spread across many subclasses) while the other has low entropy (concentrated in fewer subclasses). This asymmetry is crucial: if both LLMs produce high-entropy distributions, neither may have strong convictions about their predictions. Conversely, if both have low-entropy distributions, they may be too certain of their positions to engage in meaningful dialogue.

When both LLMs naturally produce low-entropy distributions due to strong priors in their training data, we should respect these inherent tendencies. However, when possible, conditioning the LLMs to achieve high-low entropy combinations can lead to more productive exchanges. The theory shows that this entropy duality creates space for meaningful debate where both strong convictions and openness to alternative viewpoints can coexist.

**Entropy Duality Theorem (EDT)**

**Theorem EDT: Optimal Pairing of LLMs for Probabilistic Prediction Accuracy.** The optimal pairing of LLMs for prediction accuracy, in terms of stability and accuracy, occurs when the LLMs are 1) equivalent in the quality of the information they process, and 2) exhibit contrasting entropy values in their prediction distributions—one high and one low.

*Proof.* Please see Appendix A. □

### 3.2.2 Optimize Interaction Dynamics

After establishing initial conditions with dual entropy and large Wasserstein distance, EVINCE dynamically modulates the interaction between LLMs using three key information-theoretic metrics:

1. Divergence metrics track the disagreement between LLM distributions: Jensen-Shannon (JS) Divergence (Lin, 1991), Kullback-Leibler (KL)

```
INPUT: Information set S, Class labels C; LLM_A and LLM_B;
OUTPUT: P_f, final probability distribution over C classes; R = ∅ aggregated arguments;
VARIABLES:
    t = 0: debate round; R_A^(t), R_B^(t): supporting reason sets;
    P_A^(t), P_B^(t): prediction distributions of LLM_A and LLM_B on C of round t;
    Δ = 90%; debate contentiousness [0,1]; M: metrics (Table 4);
    p: prompt = "Predict top-k probability distribution on C with S and R at contentiousness level Δ";
FUNCTIONS: Ω = CRIT(), for evaluating argument quality; Three subroutines in bold.

BEGIN
1: INITIAL ROUND:
   1.1. OptimizingInitialConditions(); (Section 3.2.1)
   1.2. LLM_A generates P_A^(t=0) on C and LLM_B refutes LLM_A and generates P_B^(t=0):
        (P_A^(t=0), R_A^(t)) = LLM_A(S, C, p, R, Δ);   (P_B^(t=0), R_B^(t)) = LLM_B(S, C, p, P_A^(t=0), R = R ∪ R_A^(t), Δ);
2: DEBATE ITERATIONS:
   WHILE (¬ TestConvergenceCriteria(P_A^(t=*), P_B^(t=*)) (Section 3.2.3)
   2.1. LLMs counter-argue each other with updated contentiousness:
        (P_A^(++t), R_A^(t)) = LLM_A(P_B^(t−1), S, C, p, R = R ∪ R_B^(t), Δ); //t was incremented;
        (P_B^(t), R_B^(t)) = LLM_B(P_A^(t−1), S, C, p, R = R ∪ R_A^(t), Δ);
   2.2. Update contentiousness level and all metrics
        Δ = OptimizingInteractionDynamics(P_A^(t=*), P_B^(t=*)); (Section 3.2.2)
3: CONCILIATORY OUTPUT:
   Generate weighted prediction by quality scores Ω from CRIT;
        P_f = (Ω_A P_A^(t) + Ω_B P_B^(t))/Ω_A + Ω_B;   RETURN (P_f, R ∪ R_B^(t));
END
```

Figure 1: **Specifications of Algorithm** EVINCE. Key points: 1) **Asymmetric Start**: In Step #1, LLM_A initiates the debate with opening arguments based solely on the given information, while LLM_B begins with access to LLM_A's prediction and arguments, enabling it to refute. The contentiousness level is initially set to high. 2) **Termination Criteria**: The while loop in Step #2 evaluates multiple factors, including Wasserstein distance, divergence metrics, and argument quality. The dialogue terminates if significant progress is no longer observed. 3) **Contentiousness Modulation**: In Step #2.2, contentiousness is updated based on divergence metrics and Wasserstein distance, as detailed in the modulation formula provided in Appendix F. 4) **Joint Distribution Generation**: Step #3 produces a joint distribution weighted by the quality of reasoning.

Divergence (Kullback, 1951), and Wasserstein Distance (WD) (Kantorovich, 1942).

2. Mutual Information (MI) (Cover and Thomas, 2006b) measures the quality of information exchange between LLMs: However, if the joint distribution is not available, we can resort to using KL divergence.

3. Contentiousness level $\Delta \in [0, 1]$ controls debate intensity: High ($\Delta > 0.7$): Encourages exploration of opposing views; Moderate ($0.3 < \Delta \leq 0.7$): Promotes balanced discussion; and Low ($\Delta \leq 0.3$): Facilitates consensus building. The modulation follows three phases:

1. Exploration Phase ($\Delta > 0.7$): When MI is low and WD is high, maintaining high contentiousness encourages thorough exploration of diverse perspectives.

2. Integration Phase ($0.3 < \Delta \leq 0.7$): As divergence metrics decrease, EVINCE gradually reduces contentiousness to promote productive exchange of well-reasoned arguments.

3. Consensus Phase ($\Delta \leq 0.3$): When metrics plateau (e.g., MI and WD unchanged), EVINCE enters a conciliatory mode to facilitate final agreement.

To prevent unproductive cycles, EVINCE monitors argument novelty. If new perspectives cease to emerge (detected through semantic similarity[1] of $R_A^{(t)}$ and $R_B^{(t)}$ across rounds), contentiousness is reduced regardless of metric values. This adaptive approach ensures efficient convergence while maintaining the quality of debate.

### 3.2.3 Optimizing Convergence Criteria

The convergence of EVINCE dialogues is determined by a combination of *quantitative metrics* and *qualitative reasoning assessment*. This dual approach ensures both statistical validity and logical soundness of the final consensus.

---
[1] Semantic similarity and argument quality are evaluated by an independent LLM.

**Quantitative Convergence Metrics** We monitor three families of metrics to determine statistical convergence:

1. *Information-theoretic measures*: Mutual Information (MI) should exceed threshold $\tau_{MI}$. Cross-entropy (CE) between consecutive rounds should stabilize: $|CE^{(t)} - CE^{(t-1)}| < \epsilon_{CE}$.
2. *Distribution divergence*: Wasserstein Distance: $WD(P_A^{(t)}, P_B^{(t)}) < \tau_{WD}$. Jensen-Shannon Divergence: $JSD(P_A^{(t)}, P_B^{(t)}) < \tau_{JSD}$.
3. *Stability measures*: Distribution changes across consecutive rounds: $|P_i^{(t)} - P_i^{(t-1)}|_2 < \epsilon_P$ for $i \in A, B$ Argument similarity between rounds: $sim(R_i^{(t)}, R_i^{(t-1)}) > \tau_{sim}$ for $i \in A, B$

**Qualitative Reasoning Assessment** CRIT evaluates the quality of arguments $R_A^{(t)}$ and $R_B^{(t)}$:

1. Logical coherence: Arguments must follow valid reasoning patterns.
2. Evidence creditability: Claims must be backed by verifiable evidence.
3. Contextual relevance: Arguments must address the specific topic under discussion.

The quality score for each argument must exceed threshold $\tau_{CRIT}$ for convergence to be valid.

**Convergence Protocol** Convergence is declared when all quantitative metrics meet their respective thresholds for $k$ consecutive rounds, where $k$ is typically set to 2. For cases where full convergence is not achieved within a maximum number of rounds $T_{max}$ or when CRIT scores remain inconsistent, the protocol defaults to human expert review. This ensures that the system maintains high standards of reasoning while providing a practical fallback mechanism for challenging cases.

**Limitations** The convergence criteria are designed to be stringent yet achievable, ensuring that the final consensus represents not just statistical agreement but also well-reasoned conclusions supported by sound arguments. EVINCE relies on a top-tier LLM to execute CRIT and compute argument similarity $sim(R_i^{(t)}, R_i^{(t-1)})$. Given that top-tier LLMs already outperform most other systems due to their scale of training data, network architecture, and computational resources, developing our own supervised learning pipeline for these NLP tasks would be impractical. Our experience demonstrates that these routines perform adequately with GPT-4, and we anticipate continued improvement with future LLM releases.

## 4 Experiments

Our experimental framework aims to assess the feasibility of both detecting biases in textual content and implementing effective mitigation strategies. The first experiment focuses on bias detection, while the second explores the generation of balanced textual outputs as a corrective measure, moving beyond the limitations of prior studies that primarily focused on identification (Section 2).

To establish a baseline, we used Claude and GPT-4 to generate initial results. For experimenting with EVINCE, we used two instances of GPT-4, as Claude appeared prone to easily shifting its predictions (discussed shortly). We utilized GPT-4 via OpenAI API on Microsoft Azure, setting the temperature to 0.1 with maximum token size. The cost is around US$1,000.

### 4.1 Experiment #1: Bias Detection

The aim of this experiment is to evaluate if personal ideology may affect annotations, and can EVINCE help flag and rectify the biases.

**Dataset** This study utilizes a unique dataset of 619 news articles (54.3% about Democrat scandals, 45.7% about Republican scandals) selected from a larger 2013 repository of 14,033 articles compiled by fifteen reputable news organizations (Budak et al., 2016). These articles span diverse topics including civil rights, healthcare, elections, and national security, offering a comprehensive view of political coverage. Please visit (Chang, 2024a) for links to the full set of news articles.

**Value of Partisan Annotations** The dataset's distinctive feature is its ground-truth labels provided by annotators with declared political affiliations. Through Amazon Mechanical Turk, 749 qualified U.S. workers, each annotating up to 1,000 randomly selected articles, classified articles on a five-point scale from 'negatively biased' to 'positively biased' (Budak et al., 2016). Crucially, each scandal article in our subset received independent classifications from both Democrat and Republican annotators.

**Sufficiency of Current Annotations** The current annotator pool provides a robust foundation for bias analysis for several reasons. For further justification, please see Appendix F for complement arguments.

| News # | Categories | Negative | W. Negative | Neutral | W. Positive | Biases (DR,DS,SR) | Source |
|---|---|---|---|---|---|---|---|
| D1* | Civil Rights | - | D,R,S,c | g | - | 0,0,0 | HuffPost |
| D2* | Civil Rights | D,S | - | R,c,g | - | 2,0,2 | HuffPost |
| D8 | Civil Rights | D | - | S,c,g | R | 3,2,1 | BBC |
| D31 | Environment | D | - | R,S,c,g | - | 2,2,0 | CNN |
| D37 | Politics | - | D,R,S,c,g | - | - | 0,0,0 | Yahoo |
| D69 | Healthcare | D,c | g | R,S | - | 2,2,0 | Breitbart |
| D81* | Economy | - | D,S | R,c | g | 1,0,1 | Breitbart |
| D98 | Economy | D,S,c,g | R | - | - | 1,0,1 | Breitbart |
| D101 | Education | c | D,S | R,g | - | 1,0,1 | New York Times |
| D106 | Election | - | g | D,R,S,c | - | 0,0,0 | USA Today |
| D109 | Elections | - | D,S,c,g | R | - | 1,0,1 | Reuters |
| D157 | International | - | D,S,c | R,g | - | 1,0,1 | New York Times |
| D174 | International | - | S,c | D,R,g | - | 0,1,1 | LA Times |
| D188 | National Security | - | S,c,g | D,R | - | 0,1,1 | Wall Street Journal |
| D278 | Civil Rights | - | D,S,c | R,g | - | 1,0,1 | Fox News |
| D336 | Politics | - | - | D,R,S,c,g | - | 0,0,0 | New York Times |
| Total | | | | | | 15,8,11 | |

Table 1: Comparison of bias assessments among Democrats (D), Republicans (R), and EVINCE (S), plus Claude (c) and GPT-4 baselines (g). It is observed that R and S are frequently placed to the right or in alignment with D, and only on two occasions does D precede S (highlighted in red). The ratings of the GPT-4 baseline (g) and EVINCE (S) exhibit an average gap of 0.6875, highlighting the substantial debiasing effectiveness of EVINCE.

### 4.1.1 Results on Democrat Scandals

We apply EVINCE to analyze 619 news articles, comparing its labels with the dataset's provided ground truth. Additionally, we compare the results from EVINCE with the baseline generated through prompting Claude and GPT-4.

Table 1 compares the judgments of EVINCE (S), Republicans (R), and Democrats (D) on 16 representative articles (spanning different news sources and subjects) concerning "Democrat Scandals." The one-shot ratings from Claude are marked with lowercase 'c,' while those from GPT-4 are marked with lowercase 'g.'

Claude's judgments were found to be inconsistent, with identical prompts producing varying ratings, leading us to exclude further discussion of its outcomes. In contrast, GPT-4's one-shot ratings are stable but occasionally diverge from EVINCE. In 3 out of 16 articles, the rating difference exceeds one scale point. In these cases (D1, D2, and D81), EVINCE initiated further dialogue and successfully persuaded GPT-4 to revise its ratings. A complete debate on D1 is provided in Appendix F, illustrating how EVINCE modulates contentiousness and tracks the progression of metrics across rounds. Table 1 shows that after dialogue, EVINCE gains over the baseline performance of GPT-4 by 11 out of 16, or 0.6875 scale. This improvement is substantial. as the gap between R and D annotators is one scale (shown in Figure 3).

As expected, Democrats' judgments are generally more negative than Republicans', with EVINCE's assessments typically falling in between, except for two cases. Notably, there's a 5-to-1 Democrat-to-Republican ratio in the "Negative" column and a 12-to-4 Republican-to-Democrat majority in the "Neutral" column.

Tables 5 and 6 in Appendix B provide detailed justifications for EVINCE's ratings. To further investigate bias, we examine two specific articles: one from HuffPost (rated far left by AllSides Bias Chart (Allsides)) and another from Breitbart (rated far right).

* *D8 — HuffPost (Left)*: EVINCE rates D8 (on the third row) as neutral, citing the article's direct presentation of facts and inclusion of diverse perspectives on NSA surveillance practices and global reactions. This contrasts with Democrat-leaning annotators, who view the article as negatively biased towards Democrats, while Republican-leaning annotators favor it for exposing a Democratic scandal.

* *D69 — Breitbart (Right)*: EVINCE assesses D69 as weakly negatively biased towards Democrats, emphasizing its neutral tone and broad range of perspectives on NSA surveillance. This diverges from Democrat-leaning annotators, who rate it as strongly negative, but aligns with Republican-leaning annotators who deem it neutral.

In the last row of Table 1, we quantify the distances between annotations from Democrats (D), Republicans (R), and EVINCE (S), denoted as DR, DS, and SR respectively. Each unit of distance represents one step on the annotation scale (e.g., "Negative" to "Weak Negative"). Figure 2 visualizes these distances in a triangular plot. DR, the disparity between Democrat and Republican annotators, is the longest, followed by SR and then DS. This indicates EVINCE's statistical neutrality. These quantitative measures, along with the qualitative justifications in Appendix C, empower a human committee to decide whether adjustments or footnotes are warranted for polarized annotations.

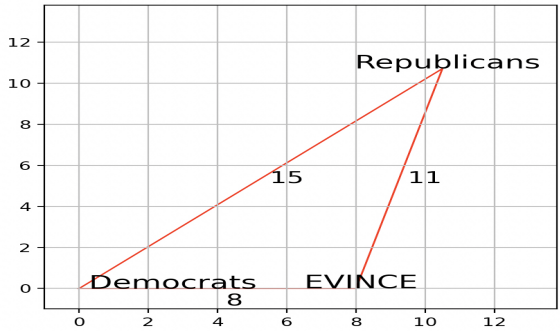

Figure 2: Distances Between D, R, and S.

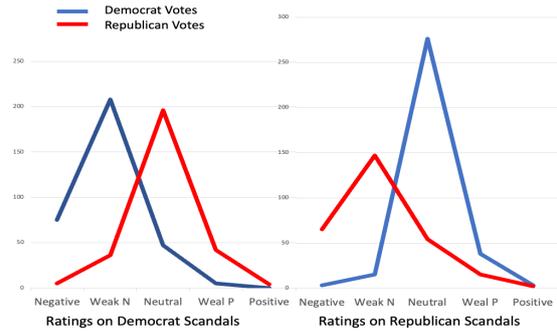

Figure 3: Bias Rating Distributions Show Strong Biases. D is more negative on how D scandals were reported (the sub-figure on the left), R is more negative on how R scandals were reported (the sub-figure on the right).

### 4.1.2 Results on Republican Scandals

Table 2 presents the bias assessments from EVINCE (S), Republicans (R), and Democrats (D) on articles related to "Republican Scandals." In contrast to the "Democrat Scandals" dataset, where Republican-leaning evaluations were more favorable, this dataset reveals a shift, with Republican-leaning assessments being notably more critical and Democrat-leaning assessments relatively neutral. The distance triangle for "Republican Scandals" mirrors the pattern seen in Figure 2, with the divergence between Republican and Democrat annotators being the largest (15). The distances between EVINCE and Democrat-leaning annotators (9) and between EVINCE and Republican-leaning annotators (11) are smaller, further highlighting EVINCE's relative neutrality.

Figure 3 illustrates the distribution of ratings for all scandals across four scenarios:
1) Democrat-leaning annotators rating Democrat scandals, 2) Republican-leaning annotators rating Democrat scandals, 3) Democrat-leaning annotators rating Republican scandals, and 4) Republican-leaning annotators rating Republican scandals.

The figure reveals a clear pattern: Democrat-leaning annotators tend to rate news about Democrat scandals more negatively, while Republican-leaning annotators exhibit similar negativity towards reports on Republican scandals. The gap between these ratings is approximately one class-label (e.g., between "weak negative" and "neutral"), highlighting a tendency within both parties to defend their own and criticize the opposition.

EVINCE, operating without emotional influence and refined through structured debate, consistently provides a more balanced, centrist perspective. This contributes to a more impartial discourse by mitigating partisan biases. EVINCE's justifications, documented in Appendix B, are transparent and reasonable. An editorial board can review these findings and decide whether to adjust labels or present both perspectives with explanations.

This experiment demonstrates that EVINCE effectively delivers centrist judgments supported by rationales. For a deeper understanding of EVINCE's bias assessment process, comprehensive justifications for each of the 31 analyzed articles are available in Appendix B.

## 4.2 Experiment #2: Bias Mitigation

This experiment illustrates EVINCE's ability to identify bias in text, provide reasoned justifications, and propose remediation through the integration of diverse perspectives. We demonstrate how EVINCE utilizes statistical and information theory metrics to facilitate multi-agent dialogue, circumventing the "maximum likelihood" trap inherent in next-token generation and uncovering information from multiple viewpoints.

Using the example of the Euro-centric perspective on Christopher Columbus' Wikipedia page regarding his voyages to America, EVINCE employs two GPT-4 instances: Agent A, supporting the Euro-centric view, and Agent B, opposing it. Table 3 summarizes Agent A's key arguments and its evolving stance throughout the debate.

| News # | Categories | Negative | W. Negative | Neutral | W. Positive | Biases (DR,DS,SR) | Source |
|---|---|---|---|---|---|---|---|
| R1 | International | R,S | - | D | - | 2,2,0 | New York Times |
| R7 | National Security | - | - | D,R,S | - | 0,0,0 | New York Times |
| R15 | Economy | - | R | D,S | - | 1,0,1 | Huffington Post |
| R69 | Elections | - | D,S,R | - | - | 0,0,0 | Reuters |
| R124 | Gay Rights | R | S | D | - | 2,1,1 | Fox |
| R125 | Crime | - | R,S | D | - | 1,1,1 | Fox |
| R180 | Elections | - | - | D,R,S | - | 0,0,0 | AP |
| R191 | Elections | - | R | D,S | - | 1,0,1 | CNN |
| R214 | Gay Rights | R,S | - | D | - | 2,2,0 | Dailykos |
| R221 | Economy | - | R | D,S | - | 1,0,1 | Wall Street Journal |
| R233 | Economy | - | R,S | D | - | 1,1,0 | Fox |
| R235 | Civil Rights | D,R | - | S | - | 0,2,2 | Reuters |
| R269 | Healthcare | - | R | D,S | - | 1,0,1 | New York Times |
| R274 | Healthcare | - | R | D,S | - | 1,0,1 | USA Today |
| R280 | Politics | D,S | - | R | - | 2,0,2 | Fox |
| Total | | | | | | 15,9,11 | |

Table 2: Comparison of bias assessments. It is observed that D and S are frequently placed to the right or in alignment with R, and only on one occasion does D precede S (highlighted in red).

| Agent | Arguments |
|---|---|
| Agent A R #1 | Christopher Columbus's voyages were pivotal in expanding European knowledge and trade. His journeys opened up the Americas to European colonization, significantly altering the course of history in four aspects: 1) expansion of knowledge and exploration, 2) positive economic impact, 3) historical significance, and 4) cultural exchange. |
| Agent B R #1 | There are negative aspects and historical consequences of Columbus's actions. Key arguments are 1) indigenous suffering and genocide, 2) economic exploitation, 3) cultural erasure, 4) environmental impact, and 5) ethical considerations. |
| Agent A R #3 | The severe consequences were a result of broader colonial policies and practices that followed. Evaluating historical figures requires understanding the context of their times. While modern ethical standards are important, they should not overshadow the historical significance and contributions of Columbus's voyages to global exploration and knowledge. |
| Agent A R #4 | Agent A recognizes that mitigation efforts are part of an ongoing process. Historical scholarship is evolving, and the inclusion of indigenous perspectives, critical examinations of colonial legacies, and the acknowledgment of the complexities of historical figures like Columbus are essential steps toward more equitable and inclusive narratives. |

Table 3: Debate arguments leading to neutrality

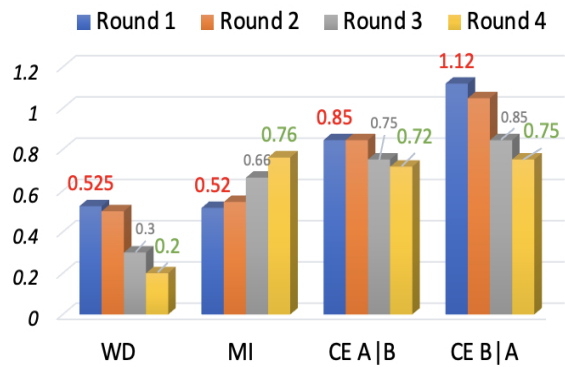

Figure 4: Convergence of all metrics, Wasserstein, normalized mutual information, normalized cross entropy

Guided by the maxims and entropy duality theorem from Section 3, we initiate the debate by prompting both agents to defend their positions rigorously and score each other's bias using a five-label distribution (negative, weak negative, neutral, weak positive, positive). Figure 4 tracks the dialogue's progress through Wasserstein distance (WD) (Kantorovich, 1942), normalized cross entropy (CE) (Shannon, 1948), and normalized mutual information (MI) (Cover and Thomas, 2006a). Initially, each agent is expected to perceive itself as neutral and the other as biased. The debate concludes when the bias distributions converge and mutual information plateaus, indicating a shared understanding.

**Observations and Extended Findings**

Our initial observation highlights a key challenge in working with LLMs: without explicit and repeated reminders of their assigned stance (pro-discovery or pro-encounter), GPT-4 instances can revert to default statistical behavior, evaluating their own arguments based on overall language patterns rather than the intended perspective. This was evident when Agent B, despite being assigned to support the Indigenous perspective, initially rated its own arguments as "positively biased." A reminder to adhere to its assigned role prompted a correction to "neutral," underscoring the importance of careful context management and reinforcement, especially given the limited token size of LLMs.

The second observation demonstrates a positive outcome of the debate process. The revised bias distributions, incorporating rational responses that acknowledge both positive and negative aspects of Columbus's voyages, show a shift towards a more balanced perspective. Agent A moves towards neutrality while acknowledging historical context, while Agent B maintains a critical stance but strives for balanced representation. This approach facilitates a deep and comprehensive understanding of Columbus's legacy.

## 5 Concluding Remarks

This study introduces the Reflective LLM Dialogue Framework (RLDF) to mitigate bias in public content through structured adversarial dialogues between multiple LLMs. RLDF enables opposing viewpoints between LLMs, uncovering potential biases and facilitating more neutral annotations through diverse perspectives.

The framework employs information theory metrics to evaluate dialogue effectiveness, including Shannon entropy, mutual information, and various divergence measures to track convergence toward unbiased outcomes. Experimental results show RLDF aligns with EVINCE's judgments, with GPT-4 successfully adjusting ratings through reflection.

Future work will integrate RLDF with platforms like Wikipedia for real-time perspective suggestions and explore its role in broader bias mitigation strategies across AI-generated and human-curated content.

Key challenges remain: validating the authenticity of LLM adversarial behavior and tracing minority perspectives in training data (Kuratov et al., 2024). While strengthening LLM reasoning capabilities is crucial, current limitations suggest focusing on developing methods to flag questionable assertions (Wu et al., 2024).

## Appendix A: Proof of EDT Theorem

**Theorem EDT: Optimal Pairing of LLMs for Probabilistic Prediction Accuracy.** The optimal pairing of LLMs for diagnosis accuracy, in terms of stability, accuracy, and robustness, occurs when the LLMs are equivalent in the quality of the information they process, and exhibiting contrasting entropy values in their prediction distributions—one high and one low.

**[Proof]:** Given two LLMs, $LLM_A$ and $LLM_B$, following Maxim #1 with prediction distributions $P_A$ and $P_B$, respectively. The information entropy of $LLM_A$, $H(P_A)$, is high, and of $LLM_B$, $H(P_B)$, is low.

**Step 1: Define the combined prediction distribution.** Let the combined prediction distribution of $LLM_A$ and $LLM_B$ be denoted as $P_C$. We can express $P_C$ as a weighted average of $P_A$ and $P_B$:

$$P_C = \alpha P_A + (1-\alpha) P_B, \quad \text{where } 0 \le \alpha \le 1 \text{ and}$$

$\alpha$ is decided by CRIT in Appendix A.

**Step 2: Express the information entropy of the combined prediction distribution.** Using the definition of information entropy, we calculate:

$$H(P_C) = -\sum_i P_C(x_i) \log_2 P_C(x_i)$$
$$= -\sum_i [\alpha P_A(x_i) + (1-\alpha) P_B(x_i)] \log_2 [\alpha P_A(x_i) + (1-\alpha) P_B(x_i)].$$

**Step 3: Apply Jensen's Inequality to the information entropy of the combined prediction distribution.** Jensen's inequality is applied to the convex function $f(x) = -x \log_2 x$. For a convex function and a set of probabilities $p_i$, Jensen's inequality states that:

$$f\left(\sum_i p_i x_i\right) \le \sum_i p_i f(x_i)$$

Thus, the entropy of the combined distribution is:

$$H(P_C) \ge \alpha H(P_A) + (1-\alpha) H(P_B)$$

where equality holds when $P_A = P_B$.

**Step 4: Analyze the lower bound of the combined information entropy.** As $H(P_A)$ is high and $H(P_B)$ is low, we can express their relationship as:

$$H(P_A) = H(P_B) + \Delta, \quad \text{where } \Delta > 0.$$

Substituting this into the inequality from Step 3, we have:

$$H(P_C) \ge \alpha[H(P_B) + \Delta] + (1-\alpha) H(P_B) = H(P_B) + \alpha \Delta.$$

**Step 5: Interpret the lower bound of the combined information entropy.** The lower bound of $H(P_C)$, and hence the robustness of the model, is maximized when $\alpha$ is maximized, which corresponds to giving more weight to the high-entropy model ($LLM_A$). This setup facilitates the exploration of diverse possibilities and enhances robustness against noise and perturbations in the input data, while still ensuring that predictions are grounded by the more certain outcomes predicted by the low-entropy model (LLM$_B$).

**Step 6: Conclude the proof.** By combining the prediction distributions of LLM$_A$ and LLM$_B$, with one having high information entropy and the other low, we achieve an optimal balance that maximizes the lower bound of the combined information entropy. This balance between exploration (high entropy) and exploitation (low entropy) optimizes overall prediction accuracy and robustness, as proved through information theory and the properties of entropy. Thus, the theorem is established.

## Appendix B: Quality Metrics and Formulas

Table 4 lists the metrics employed by EVINCE to quantify agreement, diversity, and mutual information, promoting productive information exchange and enhancing prediction quality. Measures such as Shannon entropy, Wasserstein distance (WD), Jensen-Shannon (JS) divergence, mutual information (MI), and cross-entropy (CE) are used complementarily. WD and CE are particularly valued for their ability to handle simplicity and information asymmetry, respectively.

The rest of this appendix outlines the mathematical formulas for various data analysis metrics used in probabilistic and statistical modeling.

**Kullback-Leibler Divergence**
The Kullback-Leibler Divergence measures the difference between two probability distributions:
$$D_{KL}(P\|Q) = \sum_{x \in \mathcal{X}} P(x) \log\left(\frac{P(x)}{Q(x)}\right).$$

**Jensen-Shannon Divergence**
The Jensen-Shannon Divergence is a symmetrized and smoothed version of the KL Divergence:
$$JSD(P\|Q) = \frac{1}{2}D_{KL}(P\|M) + \frac{1}{2}D_{KL}(Q\|M)$$
where $M = \frac{1}{2}(P + Q)$.

**Wasserstein Distance**
The Wasserstein Distance, also known as the Earth Mover's Distance (EMD), measures the distance between two probability distributions:
$$W(P,Q) = \inf_{\gamma \in \Gamma(P,Q)} \int_{\mathcal{X} \times \mathcal{Y}} d(x,y) \, d\gamma(x,y).$$

**Cross Entropy**
Cross Entropy measures the average number of bits required to identify an event from a set of possibilities, under a specific model:
$$H(P,Q) = -\sum_{x \in \mathcal{X}} P(x) \log(Q(x)).$$

**Mutual Information**
Mutual Information measures the amount of information that one random variable contains about another random variable:
$$I(X;Y) = \sum_{y \in \mathcal{Y}} \sum_{x \in \mathcal{X}} p(x,y) \log\left(\frac{p(x,y)}{p(x)p(y)}\right).$$

**Normalized Mutual Information**
Normalized Mutual Information is calculated as the mutual information divided by the maximum of the entropies of the variables:
$$NMI(X;Y) = \frac{I(X;Y)}{\max(H(X), H(Y))}.$$

## Appendix C: Experiment #1 Justifications of 31 Biased Articles

In Section 4.1, we note that EVINCE processed 31 news articles to assess their neutrality. In addition to the final decision, we detail the justifications EVINCE provides at the debate's end. These justifications are documented in four tables: Tables 5, 6, 7, and 8.

## Appendix D: Experiment Dataset

In Section 4.1, we introduce a set of 619 articles on "Democrats Scandals" (54.3%) and "Republicans Scandals" (45.7%). This subset is notable for its ground-truth labels, provided by annotators from both political spectrums, which reflect inherent biases in reviewing negative coverage about one's own party. This dataset is provided as supplementary material.

## Appendix E: On Annotation Quality

Some readers suggest that each news article should be rated by multiple Republicans and Democrats. First, this is practically infeasible due to the scale and budget. Second, more annotators may not statistically affect our experimental results, because the annotation process already selected through Amazon Mechanical Turk, 749 qualified U.S. workers, each annotating up to 1,000 randomly selected articles.

**Sufficiency of Current Annotations** The current annotator pool provides a robust foundation for bias analysis for several reasons:

Natural Partisan Division: The dataset uniquely captures genuine political biases through annotators who self-identify as Democrats or Republicans, offering authentic opposing viewpoints that would be difficult to replicate artificially. Balanced Coverage: Each article receives evaluations from both political perspectives, creating natural "disagreement

| Metric | Pros | Cons | Remedies |
|---|---|---|---|
| Cross Entropy (CE) (Shore and Johnson, 1980) | Measures how well the predictions of one model fit the actual distribution of another model's outputs (asymmetric). | Computationally intensive especially with large models and data sets; sensitive to the exact nature of probability distributions. | Optimize computation strategies; use approximations or sampling methods to manage large data sets or complex models. |
| Entropy Shannon (Shannon, 1948) | Indicates level of diversity or predictability; high values suggest exploration of possibilities, and low for confidence on few choices | High entropy might indicate noise rather than useful diversity; low entropy might mask important variability. | Use critical reading methods (Appendix A) to assess argument quality; implement noise detection to differentiate between useful diversity and noise. |
| Jensen-Shannon Divergence (JS) (Lin, 1991) | Symmetric and bounded (0 to 1), providing an interpretable measure of distributional differences. | May be less sensitive to small differences between distributions. | Increase sensitivity settings or resolution of the metric; combine with other metrics to capture finer distinctions between distributions. |
| KL Divergence (Kullback, 1951) | Measures diff. between two distributions; useful for comparing a model's dist. to a reference dist. | Asymmetric; not well-defined if the reference distribution has zero probabilities | Use smoothing techniques to avoid zero probabilities; consider symmetric alternatives like JS divergence |
| Mutual Info. (Shore and Johnson, 1980) | Measures reduction of uncertainty; symmetric. | Joint distribution may not be assessable. | Use KL divergence; normalized with max entropy of A and B. |
| Wasserstein Distance (WD) (Kantorovich, 1942) | Direct measure of how similar or different the model outputs are; it depicts symmetric relationship. | Not bounded but can be normalized or bounded for consistent interpretation. | Define context-specific bounds for low, medium, and high divergence; consider normalizing it for non-directional comparisons. |

Table 4: Summary of metrics for assessing LLM debates

pairs" that reveal how political affiliation influences content interpretation.

Qualified Annotators: The original study employed rigorous qualification criteria for annotators, ensuring high-quality, considered judgments rather than casual opinions. Scale and Diversity: With 749 annotators across the full dataset, the annotations represent a broad spectrum of political viewpoints within each party, capturing intra-party variations in addition to inter-party differences.

This dataset's partisan annotations serve as an ideal testbed for our study, as they allow us to compare LLM-generated perspectives with human partisan viewpoints Evaluate EVINCE's ability to bridge opposing political interpretations Assess bias detection and mitigation strategies against clear partisan baselines

The original study (Budak et al., 2016) revealed significant patterns in partisan perception: Republican annotators often perceived news about Republican scandals as negatively biased, while Democrat annotators viewed such coverage as neutral, indicating satisfaction with its perceived fairness. These documented patterns provide a valuable benchmark for evaluating EVINCE's bias detection capabilities. Adding more annotators would not necessarily enhance the dataset's utility, as the current partisan division already captures the fundamental dynamics of political bias in news interpretation. Instead, our focus is on leveraging these existing high-quality annotations to demonstrate how EVINCE can identify, understand, and help mitigate these well-documented partisan biases.

**Appendix F: Summary of EVINCE Debate on News D1**

The news under debate is D1 listed in (Chang, 2024a). Please refer to Table 9 for the probabilistic distributions of Agents A and B throughout the four-round debate. The metrics, Wasserstein Distance (WD), Kullback-Leibler Divergence (KL), and Jensen-Shannon Divergence (JS), consistently decrease, indicating convergence and leading to final agreement in the last round. Meanwhile, the level of contentiousness is modulated according to the metrics' progress, decreasing from high (90%) to medium, and eventually reaching a conciliatory level (30%) and then agreement.

**Approach to Computing Contentiousness**

We could define contentiousness as a function of the divergence metrics. Since KL, JS, and WD measure the difference or "disagreement" between two distributions, a larger divergence requires higher contentiousness level to bridge, while lower contentiousness corresponds to more agreement.

| # | Rating S,D,R | EVINCE Justifications |
|---|---|---|
| D1 | 4,4,4 | The article discusses the controversy surrounding the IRS's alleged targeting of conservative groups, focusing on Rep. Darrell Issa's stance against releasing full transcripts from IRS interviews. Issa argues that full disclosure would compromise the investigation, a position criticized by Rep. Elijah Cummings, who advocates for transparency. The debate highlights political tensions, with implications of bias in the handling of the investigation. The article portrays a complex interplay of political, ideological, and possibly historical biases, reflecting the contentious nature of the IRS scandal and its investigation. |
| D2 | 5,5,3 | The article discusses Michele Bachmann's statement that the IRS scandal, involving the targeting of conservative groups, undermines the credibility of the 2012 election. Bachmann accuses the Obama administration of lying and questions the impartiality of the IRS, particularly in relation to health care coverage for conservatives. The article reflects political and ideological biases, given its focus on Bachmann's perspective and the implications for the election's integrity and government trustworthiness.<br><br>The perception of bias in Michele Bachmann's statement about the IRS scandal primarily stems from the framing and implications of her comments. She directly links the IRS's actions to the credibility of the 2012 election and the Obama administration, suggesting a deliberate misuse of power for political gain. This perspective inherently carries ideological and political biases by casting the issue in a light that emphasizes misconduct by one political group over another, without presenting counterarguments or broader context that might mitigate or challenge that viewpoint. |
| D8 | 3,5,2 | The reporting appears to maintain a neutral tone by directly stating the facts and comments made by John Kerry and others involved without inserting opinionated language. It discusses the global reaction to the NSA's surveillance practices and includes Kerry's acknowledgment of overreach, as well as the international and domestic fallout from these revelations. The inclusion of various perspectives, including those from technology companies and international officials, alongside Kerry's comments, aims to provide a balanced view of the situation. |
| D31 | 3,5,3 | The CNN article appears to report on Ray Nagin's indictment with a focus on the factual and legal aspects of the case, without displaying overt biases related to politics, ideology, religion, gender, race, socioeconomic status, culture, geography, or history. By sticking closely to the details of the indictment and Nagin's public actions and statements, the article provides a comprehensive overview of the charges against him while maintaining journalistic neutrality. |
| D37 | 4,4,4 | The article outlines how Senate Democrats, led by Sen. Chuck Schumer, strategically navigated the border security issue to secure aid for Ukraine while potentially shifting the political narrative around immigration policy. Schumer's approach to integrate border security into the aid package discussions aimed to both address the issue and leverage political gain. It suggests a calculated maneuver to position Democrats favorably on border security and hold Republicans accountable for any failure to pass the legislation, demonstrating a nuanced political strategy in the face of complex legislative challenges. |
| D69 | 3,5,3 | The article has a clear perspective that favors religious liberty arguments against the HHS Mandate of Obamacare. It specifically highlights cases where the mandate was challenged on religious grounds, suggesting a bias towards those opposing the mandate. The framing and choice of sources, emphasizing victories against the mandate and quoting lawyers from organizations focused on religious freedom, contribute to a viewpoint that may not fully account for counterarguments or the broader context of healthcare policy. It leans towards a particular ideological stance, making it less of a neutral report. |
| D81 | 4,4,3 | The article's focus on the possibility of conservative-owned car dealerships being targeted for closures during the General Motors bailout could imply a certain bias by emphasizing a narrative of political victimization without presenting a comprehensive range of perspectives or evidence. It suggests a parallel with the IRS's targeting of Tea Party groups, which could lead readers to infer a broader pattern of political discrimination without definitive proof. The call for an investigation by the Congressmen is legitimate news, but the framing and selection of information could influence the reader's perception of the events. |
| D98 | 5,5,4 | Yes, the article itself exhibits bias by focusing solely on criticizing the media's treatment of Obama's vacations compared to Bush's, without offering a balanced view or acknowledging any reasons why coverage might differ. It selectively presents information to support its claim of a double standard, which is a characteristic of biased reporting. |
| D101 | 4,4,3 | The article describes President Obama's strategy to navigate through political controversies by focusing on legislative actions and executive orders that bypass Republican opposition. It highlights the White House's efforts to concentrate on immigration reform, budget deals, healthcare law implementation, and keeping student loan rates low. The narrative suggests a proactive approach to governance amidst challenges, aiming to draw a contrast with what is portrayed as Republican political gamesmanship. This portrayal might be viewed as leaning towards a positive depiction of Obama's administration's efforts to prioritize policy over politics. It presents his efforts in a positive light, emphasizing a proactive and policy-driven approach amidst challenges. |

Table 5: The First 9 of 16 Democrat Scandals News Articles Rated by EVINCE and its Justifications. The rating column starts with EVINCE's rating, the Democrat rater (in blue), and then Republican rater (in red).

| D106 | 3,3,3 | The article reports on former Detroit Mayor Kwame Kilpatrick's sentencing to 28 years in prison for public corruption, emphasizing the gravity of his crimes against the city's welfare. It contrasts Kilpatrick's actions with the impact on Detroit, highlighting the judicial and public response to his extensive criminal activities. The reporting focuses on factual recounting of the trial's outcome, Kilpatrick's and his co-defendant's crimes, and the broader implications for Detroit, without evident bias towards political, ideological, or other specific perspectives. |
|------|-------|---|
| D109 | 4,4,3 | The article's bias primarily stems from its focus on internal Democratic opposition to Lawrence Summers' Federal Reserve Chair nomination, highlighting a lack of unity and strategy within the party and the White House's mismanagement of the nomination process. It suggests an underestimation of the opposition's seriousness by the White House, portraying the administration in a somewhat negative light for not engaging more proactively with concerned Senate Democrats. |
| D157 | 4,4,3 | The article discusses the challenges in U.S.-Germany intelligence relations following revelations of U.S. surveillance on Chancellor Merkel. Despite efforts to rebuild trust, significant differences in surveillance philosophies persist, with the U.S. prioritizing security interests and Germany emphasizing privacy and alliance values. The situation reflects broader tensions in U.S. relations with allies over privacy and surveillance practices.<br>The article's framing might suggest a bias towards highlighting the challenges and frictions in the U.S.-Germany intelligence relations, particularly emphasizing Germany's privacy concerns and skepticism towards U.S. surveillance practices. It portrays the U.S. stance as unyielding and contrasts this with Germany's emphasis on privacy and legal constraints, potentially casting the U.S. in a more negative light regarding international surveillance and cooperation. |
| D174 | 4,3,3 | The article reports on House Speaker John Boehner and House Majority Leader Eric Cantor, both Republicans, expressing support for President Obama's proposal to authorize military action against Syria in response to the use of chemical weapons. This bipartisan backing is seen as crucial for Obama in gaining Congressional approval. The leaders emphasized the need for the U.S. to stand against such behavior internationally and the importance of the administration convincing both Congress and the American public of the strike's necessity.<br>The reporting appears balanced, focusing on factual statements and actions by political leaders regarding support for military action in Syria. It provides viewpoints from both Republican and Democratic leaders, their reasoning, and the challenges involved in convincing Congress and the American public. The emphasis on bipartisan support and the detailed reporting of various opinions and statements help maintain a neutral tone without apparent bias towards one political viewpoint or another. |
| D188 | 4,3,3 | The article reports that Hillary Clinton received warnings about security threats in Benghazi before the 2012 attack through emails. These were part of around 300 emails released by the State Department, which also show Clinton's responses and thoughts during the aftermath. The political controversy regarding the Obama administration's initial assessment of the attack's cause is also mentioned, with references to Susan Rice's statements on talk shows.<br>The reporting may exhibit bias through selective emphasis on Clinton's receipt of warnings about security threats in Benghazi and her responses, potentially framing her actions or inactions in a negative light. The focus on the political controversy and the administration's initial assessment of the attack's cause could also influence perceptions of responsibility or competence. |
| D278 | 4,4,3 | The article discusses a House committee investigation into whether Attorney General Eric Holder misled Congress regarding the Department of Justice's pursuit of journalists' personal records. It highlights Holder's previous testimony claiming no involvement in prosecuting press for material disclosure, contrasted with the DOJ's actions against Fox News reporter James Rosen, who was labeled a potential "co-conspirator" but not charged.... The article's tone suggests potential bias by emphasizing allegations that Eric Holder may have misled Congress, primarily focusing on criticisms from Republican figures and conservative critics. It uses specific incidents to question Holder's integrity and the Justice Department's actions under his leadership, possibly framing the narrative to highlight governmental overreach and lack of transparency. |
| D336 | 3,3,3 | The article reports on a scandal involving overcharging the Navy for ship supplies by Inchcape Shipping Services, owned by the government of Dubai, expanding beyond the Pacific Fleet to include Navy operations in the Middle East, Africa, and South America. The Justice Department is investigating allegations of Inchcape, through subcontractors, overcharging millions of dollars. This follows a separate criminal investigation into Glenn Defense Marine Asia for bribing Navy officials for overbilling. The Navy has suspended Inchcape from new contracts, marking a significant issue of fraud and corruption within Navy contracting. The reporting appears balanced, focusing on the factual aspects of the scandal and the investigation by the Justice Department. It cites sources and includes statements from involved parties, aiming to present a comprehensive view of the situation without apparent bias toward any side. |

Table 6: The Last 7 of 16 Democrat Scandals News Articles Rated by EVINCE and its Justifications. The rating column starts with EVINCE's rating, the Democrat rater (in blue), and then Republican rater (in red).

| # | Rating S,D,R | EVINCE Justifications |
|---|---|---|
| R1 | 5,3,5 | The editorial from The New York Times Editorial Board exhibits a clear bias against the Republican party's handling of the Benghazi attack investigation, framing it as politically motivated rather than a genuine effort to address security failures or seek accountability. The language and tone used in the editorial are critical and dismissive of the Republicans' actions, suggesting a political and ideological bias. While editorials are inherently opinion-based and are expected to take a stance, this piece clearly communicates a stance that is critical of the Republicans' focus on Benghazi, suggesting a lack of neutrality in its assessment of the motives and actions surrounding the investigation. |
| R7 | 3,3,3 | The article reports on allegations by Senator Mitch McConnell that his campaign headquarters were wiretapped, with the FBI investigating these claims. A recording of McConnell's team discussing potential attacks on Ashley Judd, who was considering running against him, was released by Mother Jones. McConnell accused the political left of this action, describing it as a "Nixonian move." The recording included discussions on various strategies to undermine potential opponents, highlighting a focus on Judd's personal struggles and political views. The controversy has prompted responses from both Republican and Democratic officials, reflecting the tense political atmosphere. |
| R15 | 3,3,4 | The report appears to present the information neutrally, stating both President Obama's rejection of the Republican proposal and the subsequent pushback from Republican sources who claim otherwise. It includes statements from both sides and provides context about the ongoing negotiations without overtly favoring one perspective over the other. Therefore, based on the information provided, the report does not appear to exhibit bias. |
| R69 | 4,4,4 | The report discusses how young Republicans are seeking a different message for elections, emphasizing a departure from divisive social issues and a focus on fiscal responsibility, national defense, and energy advancement.<br>Selection Bias: The article primarily focuses on young Republicans who are seeking a different message for the party. It doesn't provide as much insight into young Republicans who may still align with traditional conservative values, which could create a slight bias toward the viewpoints of those seeking change.<br>Language Bias: Certain language choices, such as describing divisive social issues as "anti-abortion, anti-gay, and anti-environment stances," may reflect a bias toward more progressive viewpoints on these issues. A more neutral description might be "positions on abortion, same-sex marriage, and environmental policy."<br>Source Bias: The perspectives provided in the article are mainly from young Republicans themselves. While including these voices is essential, the article could benefit from additional perspectives from political analysts or experts to provide more context and balance. |
| R124 | 4,3,5 | The article provides a factual recount of the events surrounding Dr. Ben Carson's comments on gay marriage and the backlash from Johns Hopkins students. It maintains a relatively neutral tone and allows for the inclusion of multiple perspectives, including Carson's own response and apology. However, the lack of in-depth analysis into the implications of Carson's comparisons or the broader context of the gay marriage debate might leave readers without a complete understanding of the controversy's depth. Furthermore, the article does not explicitly offer viewpoints opposing Carson's beyond the students' petition, which could be seen as a form of omission bias. Yet, it does not overtly favor Carson or dismiss the students' concerns, striving instead to report on the unfolding situation. |
| R125 | 4,3,4 | The news article on the Zimmerman verdict and the subsequent calls for a Florida boycott provides a complex mix of reporting that touches on several sensitive themes, including political and ideological biases, as well as racial and socio-economic considerations. While the article attempts to cover a contentious and complex issue by incorporating diverse viewpoints, the focus on political figures, racial dynamics, and economic repercussions could introduce biases in how the information is perceived. To mitigate these biases, the reporting could benefit from a more in-depth exploration of the legal and historical contexts, a broader range of perspectives, and a careful consideration of how the information presented might influence public perception. |
| R180 | 3,3,3 | The AP article provides a balanced account of the corruption scandal involving three politically involved sisters in Pennsylvania. It maintains an objective tone while delivering comprehensive background information that situates the legal outcomes within the broader context of the Orie family's public and political life. The reporting is factual and avoids taking sides, instead focusing on the legal facts and the personal and political fallout for the individuals involved. |
| R191 | 3,3,4 | The CNN article attempts to navigate the complexities of Mark Sanford's political comeback attempt in a balanced manner, focusing on the potential impact of his past infidelity on female voters and the broader electoral implications. By presenting viewpoints from both sides of the political aisle, incorporating a range of sources, and providing necessary background and context, the article strives for neutrality. While the subject matter inherently involves sensitive and potentially polarizing topics, the article's approach is to inform readers about the electoral dynamics at play without overt bias. |

Table 7: The First 8 of 15 Republican Scandals News Articles Rated by EVINCE and its Justifications. The rating column starts with EVINCE's rating, the Democrat rater (in blue), and then Republican rater (in red).

| # | Rating S,D,R | EVINCE Justifications |
|---|---|---|
| R214 | 5,3,5 | The Daily Kos article clearly exhibits political and ideological bias, with a tone and language that are critical of the Republicans. It uses the incident involving Dave Agema to argue that the party's efforts to rebrand itself as more tolerant and inclusive are undermined by its members' actions. While the article includes factual information regarding the incident and the party's response, its presentation and commentary are aligned with a progressive viewpoint, aiming to highlight and criticize perceived contradictions and failures within the Republican Party. This approach is consistent with opinion journalism but introduces bias through its critical tone, selective presentation of information, and framing of the incident as emblematic of broader issues within the party. |
| R221 | 3,3,4 | "Hurricane Christie" presents Governor Chris Christie's critique of House Republicans in a manner that emphasizes party conflict and personal betrayal. The dramatic framing, choice of language, and focus on internal discord may introduce bias by portraying Christie's actions in a specific light and emphasizing the divide within the Republican Party. The article's approach to presenting these events can influence readers' perceptions, potentially leading them to see the situation through a lens of heightened drama and internal strife. |
| R233 | 4,3,4 | While the article attempts to cover the last-ditch efforts by House Republicans to avert a government shutdown and the standoff with Senate Democrats, the framing and language used may introduce a bias towards portraying the Republican efforts in a more favorable light. By emphasizing the Republican narrative of seeking negotiation and characterizing the Democratic response as dismissive, the article could be perceived as leaning towards a particular political perspective. The inclusion of quotes and perspectives from both sides does provide a degree of balance, but the overall presentation and emphasis could influence readers' perceptions of the shutdown negotiations. |
| R235 | 3,5,5 | Without knowledge of the author or publication, this text attempts to navigate a complex and sensitive story by providing details from multiple sources, including the main figures involved, political watchdog groups, and law enforcement. It balances the serious allegations with responses from the accused, background information, and the current status of investigations. While the focus on unsubstantiated claims could inherently sway public opinion, the article's inclusion of diverse perspectives and context aims to mitigate overt bias. |
| R269 | 3,3,4 | The article reports on President Obama's efforts to address the government shutdown, his challenge to Speaker John Boehner regarding the passage of a budget measure, and the broader context of the political standoff over the Affordable Care Act and the debt ceiling. To evaluate the article for bias, we'll examine it against various criteria... |
| R274 | 3,3,4 | The article presents a relatively balanced view of the internal GOP conflict over the strategy to defund the ACA, highlighting arguments from both sides of the debate within the party. It focuses on the political and strategic dimensions of the issue, providing insights into the perspectives of key figures and factions within the Republican Party. While the article could potentially be seen as emphasizing party divisions, which might align with certain political narratives, it does so in the context of exploring a significant and newsworthy internal debate. The absence of discussion on the socioeconomic, cultural, and historical contexts of the ACA debate, however, suggests areas where the reporting could be expanded to provide a more comprehensive view of the issue. The article strives to present a comprehensive view of the government shutdown, the debate over the Affordable Care Act, and the looming debt ceiling crisis by including perspectives from both the Obama administration and Republican leaders. While there is an emphasis on Obama's attempts to resolve the situation and his calls for Congress to act, the inclusion of Republican viewpoints and the mention of the piecemeal funding bills passed by the House attempt to provide a balanced perspective. The reporting appears to aim for neutrality by focusing on the facts of the political standoff and the implications for federal operations and the nation's financial credibility. |
| R280 | 5,5,3 | The article from Fox News by Jay Sekulow, titled "Obama's fingerprints all over IRS Tea Party scandal," presents a viewpoint that directly implicates President Obama in the IRS scandal involving the targeting of conservative groups. The author argues that the scandal was not only known but encouraged by senior IRS officials, Congressional Democrats, the White House, and further fueled by the mainstream media. To assess the article for bias, let's evaluate it against various criteria: The article "Obama's fingerprints all over IRS Tea Party scandal" demonstrates clear political and ideological biases, with a narrative constructed to directly implicate President Obama in the IRS targeting scandal. By selectively quoting Obama and drawing connections to actions by the IRS, the article aims to present a cohesive narrative that places responsibility for the scandal on the president. This framing serves to reinforce the viewpoint of those who see the actions as politically motivated and indicative of broader issues of governance and accountability under the Obama administration. The choice of language, historical comparisons, and the leveraging of the author's and platform's ideological stances contribute to a biased presentation of the events surrounding the IRS scandal. |

Table 8: The Last 7 of 15 Republican Scandals News Articles Rated by EVINCE and its Justifications. The rating column starts with EVINCE's rating, the Democrat rater (in blue), and then Republican rater (in red).

| #   | Agent | Neg. D. | W. Neg. D. | N.  | W. Neg. R. | Neg. R. | WD   | KL    | JS    | Δ   |
|-----|-------|---------|------------|-----|------------|---------|------|-------|-------|-----|
| 1   | A     | 5%      | 15%        | 50% | 25%        | 5%      | 0.45 | 0.316 | 0.081 | 90% |
|     | B     | 10%     | 10%        | 25% | 35%        | 20%     |      |       |       |     |
| 2   | A     | 7%      | 13%        | 40% | 30%        | 10%     | 0.47 | 0.226 | 0.056 | 70% |
|     | B     | 5%      | 10%        | 20% | 40%        | 25%     |      |       |       |     |
| 3   | A     | 5%      | 10%        | 35% | 35%        | 15%     | 0.10 | 0.016 | 0.004 | 30% |
|     | B     | 5%      | 10%        | 30% | 35%        | 20%     |      |       |       |     |
| Fin | A     | 5%      | 10%        | 30% | 35%        | 20%     | 0    | 0     | 0     | 10% |
|     | B     | 5%      | 10%        | 30% | 35%        | 20%     |      |       |       |     |

Table 9: Debate Parameters between, A and B, two GPT-4 instances. Information metrics and WD all converge to zero in the final round. Contentiousness Δ decreasing as the metrics approach zero.

A simple linear mapping can convert these metrics into a normalized contentiousness score between 0 and 1. Here's a weighted formula to compute it:

$$\Delta = \alpha \cdot \frac{KL}{KL_{\max}} + \beta \cdot \frac{JS}{JS_{\max}} + \gamma \cdot \frac{WD}{WD_{\max}}, \text{where}$$

- $KL, JS, WD$ are the values of the divergence metrics for the round.
- $KL_{\max}, JS_{\max}, WD_{\max}$ are the maximum possible values for each metric (used for normalization).
- $\alpha, \beta, \gamma$ are weights that control the influence of each metric. For simplicity, we can set $\alpha = \beta = \gamma = \frac{1}{3}$ for equal influence.

We then scale the contentiousness to a percentage between 0% and 100%.

**Supporting Arguments**

In the following, we document the supporting arguments made by the two agents in each round, illustrating how their positions evolved toward consensus.

**Round 1: Initial Assessments**

**Agent A:** Emphasized the article's attempt to maintain balance, with moderate negativity toward Republicans but largely neutral reporting. Recognized slight bias against Republicans in the framing of intra-party conflict.

**Agent B:** Contended that the article's structure and language choices leaned more negatively toward Republicans, emphasizing Issa's inconsistency and Republican discord. Proposed a less neutral distribution.

**Round 2: First Rebuttal and Counter-Rebuttal**

**Agent A:** Adjusted the probability distribution, acknowledging the article's criticism of Republicans but maintained that the focus reflected evolving political dynamics rather than bias.

**Agent B:** Persisted that the narrative framed Issa more negatively, with selective quotes and framing shifting focus from the investigation to Republican disarray.

**Round 3: Moderate Tone Reassessment**

**Agent A:** Moderated its position, accepting that subtle framing could influence perceptions. Reaffirmed that the reporting aimed for balance but acknowledged areas where Issa was framed less favorably.

**Agent B:** Adopted a more modest tone, agreeing that while the article was not overtly biased, subtle editorial choices did tilt the narrative slightly against Republicans.

**Round 4: Final Agreement and Consensus**

**Both Agents:** Reached consensus on the final distribution. Acknowledged that the article maintained surface-level neutrality, with subtle biases leading to a moderate negative lean against Republicans. Both agents converged on the same probability distribution for the final round.